\documentclass[10pt,twocolumn,letterpaper]{article}
\usepackage{arxiv_wacv}
\usepackage{mystyle}
\usepackage[toc,page]{appendix}

\def\wacvPaperID{393} 

\wacvfinalcopy 

\ifwacvfinal
\usepackage[breaklinks=true,bookmarks=false]{hyperref}
\else
\usepackage[pagebackref=true,breaklinks=true,colorlinks,bookmarks=false]{hyperref}
\fi

\begin{document}

\title{Improving Object Detection by Label Assignment Distillation}

\author{Chuong H. Nguyen\thanks{Equal Contribution.}, Thuy C. Nguyen\footnotemark[1], Tuan N. Tang, Nam L.H. Phan \\
CyberCore AI, Ho Chi Minh, Viet Nam \\
{\tt\small {chuong.nguyen,thuy.nguyen,tuan.tang,nam.phan}@cybercore.co.jp}
}

\maketitle

\begin{abstract}
Label assignment in object detection aims to assign targets, foreground or background, to sampled regions in an image. Unlike labeling for image classification, this problem is not well defined due to the object's bounding box. In this paper, we investigate the problem from a perspective of distillation, hence we call Label Assignment Distillation (LAD). 
Our initial motivation is very simple, we use a teacher network to generate labels for the student. This can be achieved in two ways: either using the teacher's prediction as the direct targets (soft label), or through the hard labels dynamically assigned by the teacher (LAD). Our experiments reveal that: (i) LAD is more effective than soft-label, but they are complementary. (ii) Using LAD, a smaller teacher can also improve a larger student significantly, while soft-label can't.
We then introduce Co-learning LAD, in which two networks simultaneously learn from scratch and the role of teacher and student are dynamically interchanged. Using PAA-ResNet50 as a teacher, our LAD techniques can improve detectors PAA-ResNet101 and PAA-ResNeXt101 to $46 \rm AP$ and $47.5\rm AP$ on the COCO test-dev set. With a stronger teacher PAA-SwinB, we improve the students PAA-ResNet50 to $43.7\rm AP$ by only \1x schedule training and standard setting, and PAA-ResNet101 to $47.9\rm AP$, significantly surpassing the current methods. Our source code and checkpoints are released at \href{https://github.com/cybercore-co-ltd/CoLAD}{https://git.io/JrDZo}.
\end{abstract}

\section{Introduction}
Object detection is a long-standing and fundamental problem, and many algorithms have been proposed to improve the benchmark accuracy. Nevertheless, the principal framework is still unchanged: an image is divided into many small sample regions, each is assigned a target, on which the detector is trained in a supervised manner. Reviewing the literature, a majority are dedicated to inventing new architectures \cite{liu2016ssd,Redmon2016, redmon2018yolov3, bochkovskiy2020yolov4, tian2019fcos,law2018cornernet,ren2015faster,cai2018cascade} or defining effective training loss functions \cite{lin2017focal, cao2020prime,li2020generalized,
	rao2018learning}. However, the most frontal problem of supervised learning, \ie, how to assign the training targets, yet get less attention.

Unlike image classification, where a category can be easily set to an image, defining labels in object detection is ambiguous due to bounding boxes' overlapping. Obviously, if a region is completely overlapping or disjoint with a ground truth box, it is definitely positive (foreground) or negative (background). However, for a partial overlapping case, how should we consider it?

\begin{figure}[t]
	\centering
	\begin{subfigure}{\columnwidth}
		\centering
		\includegraphics[width=0.75\columnwidth]{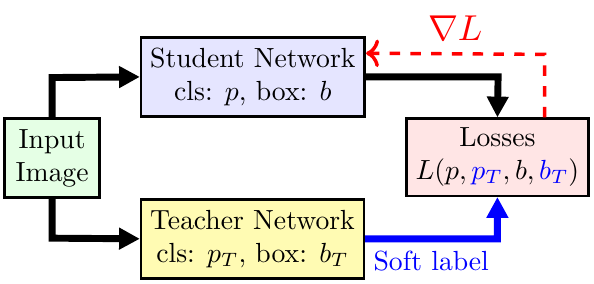}
		\caption{Soft Label Distillation}
		\label{fig:Soft-Label}
	\end{subfigure}%
	\\
	\begin{subfigure}{\columnwidth}
		\centering
		\includegraphics[width=0.95\columnwidth]{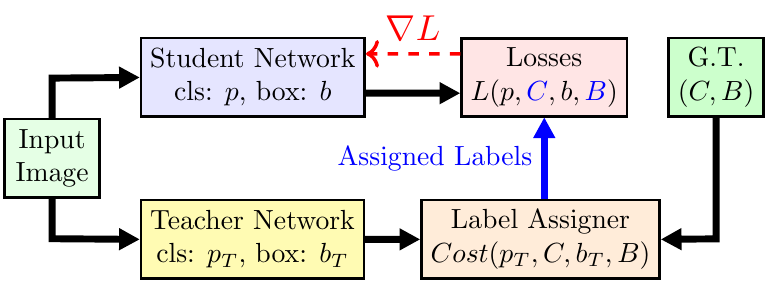}
		\caption{Label Assignment Distillation}
		\label{fig:LAD}
	\end{subfigure}
	\caption{Compare Soft Label Distillation and Label Assignment Distillation (LAD). In the former, the teacher's output is directly used as the target, while in the latter, it is used to evaluate the cost for label assignment.}
	\label{fig:Motivation}
\end{figure}

\begin{figure*}[!th]
	\centering
	\begin{subfigure}{.49\textwidth}
		\centering
		\includegraphics[width=\linewidth]{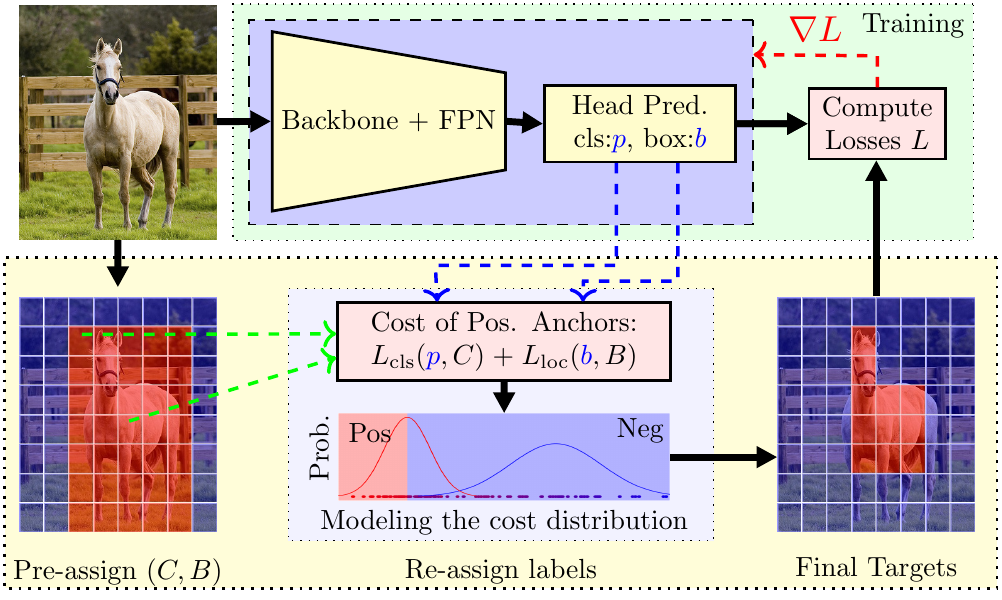}
		\caption{Probabilistic Anchor Assignment (PAA)}
		\label{fig:PAA_diag}
	\end{subfigure}%
	~
	\begin{subfigure}{.49\textwidth}
		\centering
		\includegraphics[width=\linewidth]{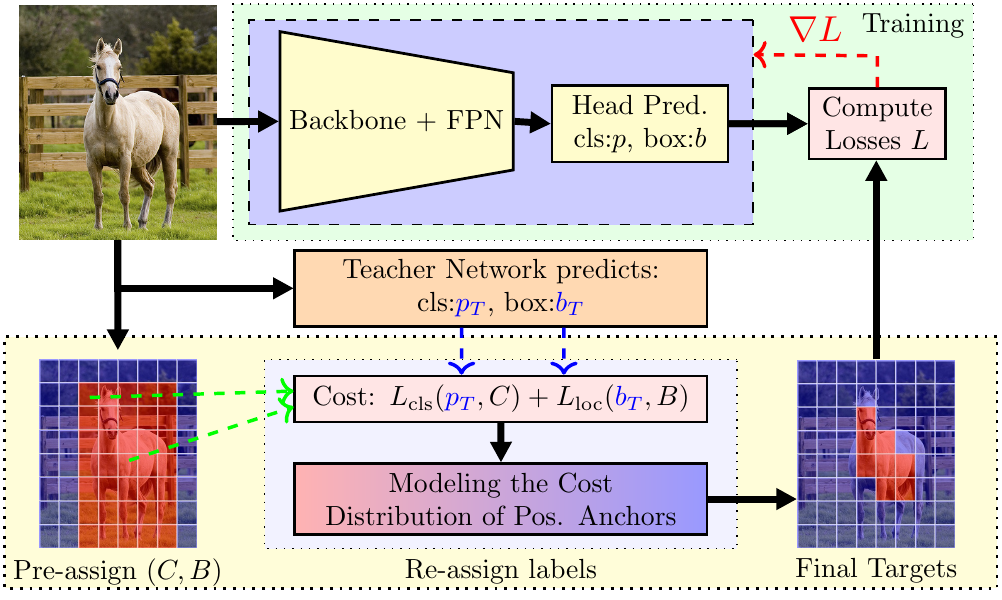}
		\caption{Label Assignment Distillation (LAD)}
		\label{fig:LAD_diag}
	\end{subfigure}
	\caption{PAA\cite{kim2020probabilistic} and its LAD counterpart. PAA uses the prediction at step $(t)$ to compute the label assignment cost for the next step $(t\!+\!1)$. It is a type of bootstrap learning, where the network updating and label assignment form a closed-loop system. In contrast, LAD uses an independent teacher, which decouples the label assignment and training processes.}
\end{figure*}

A common practice to define a positive region is if its Intersection over Union (IoU) with the nearest ground truth box at least 0.5. This may due to the high recall preference in the VOC \cite{VOC} and COCO \cite{COCO} evaluations. Ironically, a network trained with low IoU assignment yields high recall but noisy predictions, while using only high IoU samples degrades the performance \cite{Cai2019}. In addition, regions with the same IoU can have different semantic meaning for classification. Hence, recent research \cite{kim2020probabilistic, autoassign} suggests that solely relying on IoU is insufficient, and a combination of classification and localization performance is preferred. 


This paper investigates the label assignment problem from the perspective of distillation technique, hence we call Label Assignment Distillation (LAD). Regarding network distillation, the common approach is to directly use the teacher output as the targets, i.e. dark knowledge \cite{hinton2015distilling}, to train the student as illustrated in \fig{Soft-Label}. Note that, for object detection, the classification and localization tasks are often distilled independently, with or without using the ground truth. In contrary, \fig{LAD} illustrates our proposed LAD, where the classification and localization prediction of the teacher and the ground truth are fused into the cost function before being indirectly distilled to the student. 

Our proposed LAD is inspired from the realization that the label assignment methods  \cite{Carion2020, kim2020probabilistic, autoassign, ge2021ota} are forms of self-distillation \cite{caron2021emerging, grill2020bootstrap, liu2020metadistiller}. For instance, \fig{PAA_diag} illustrates the Probabilistic Anchor Assignment (PAA) \cite{kim2020probabilistic}. As shown \fig{LAD}, from LAD perspective the network itself can be seen as the teacher, that is, the current prediction is used to assign label for the next learning step. In contrary, \fig{LAD_diag} illustrates the LAD counterpart, where an independent teacher is used to compute the assignment cost, hence decoupling the training and label assignment processes.   

Motivated by this observation, we conduct a pilot study about LAD's properties. Consequently, we propose the Co-learning LAD, in which both networks can be trained from scratch, and the roles of teacher and student are dynamically interchanged. Our contributions are summarized as:
\begin{itemize}
	\item We introduce the general concept of Label Assignment Distillation (LAD). LAD is very simple but effective, applicable to many label assignment methods, and can complement other distillation techniques. More important, using LAD, a smaller teacher can improve its student significantly, possibly surpassing its teacher.
	\item We propose the Co-learning Label Assignment Distillation (CoLAD) to train both student and teacher simultaneously. We show that two networks trained with CoLAD are significantly better than if each was trained individually, given the same initialization.	
	\item We achieve state-of-the-art performance on the MS COCO benchmark. Using PAA-ResNet50 as a teacher, CoLAD improves PAA-ResNet101 to $46.0\rm AP (+1.2)$, and PAA-ResNeXt101 to $47.5\rm AP(+0.9)$. With Swin-B Transformer backbone, PAA detector trained with LAD achieves $51.4\rm AP$ on \textit{val} set, approaching the Cascade Mask-RCNN ($51.9\rm AP$) \cite{liu2021Swin}, without the costly segmentation training. Remarkably, the results are all achieved simply with smaller teachers. Finally, recycling the PAA-Swin-B as a teacher, we improve PAA-ResNet50 to $43.7\rm AP$ (with \1x schedule training), and PAA-ResNet101 to $47.9\rm AP$.
\end{itemize}

\section{Related Work}
\subsection{Label Assignment in Object Detection}
\label{sec:objdetection}
\begin{table*}[!t]
	\centering
	\caption{Comparing hard assignment methods in three aspects: pre-assignment, cost evaluation, and re-assignment.}
	\label{tab:HardLearningCompare}
	\small{
		\begin{tabular}{|l|l|l|l|}
			\hline
			Methods & Pre-assignment                  & Cost Evaluation & Re-assignment                                                                                                                                          \\ \hline
			ATSS \cite{Zhang2020}    & \begin{tabular}[c]{@{}l@{}}Top-k points closest  \\ to the object's center \end{tabular} & Anchor IoU         & \begin{tabular}[c]{@{}l@{}}Modeling the scores by a Gaussian.\\ Select the mean plus standard deviation as threshold.\end{tabular}                \\ \hline
			MAL \cite{Ke2020}     & Anchor IoU \textgreater 0.5 & \begin{tabular}[c]{@{}l@{}} Classification and \\ Localization loss\end{tabular} & \begin{tabular}[c]{@{}l@{}}All-to-top 1. The number of positive anchors gradually\\ reduces from all (first iters) to top-1 (last iters).\end{tabular} \\ \hline
			PAA\cite{kim2020probabilistic}     & Anchor IoU \textgreater 0.1 & \begin{tabular}[c]{@{}l@{}} Classification and \\ Localization loss\end{tabular} & \begin{tabular}[c]{@{}l@{}}Modeling the scores by a mixture of two Gaussian. \\ Select center of the Gaussian with lower mean as threshold.\end{tabular}                    \\ \hline
			DETR \cite{Carion2020}   & All points                  & \begin{tabular}[c]{@{}l@{}} Classification and \\ Localization loss\end{tabular}  & \begin{tabular}[c]{@{}l@{}}Modeling the scores as an optimal transportation cost. \\ Using Hungarian assignment to select top-1.\end{tabular}         \\ \hline
			OTA \cite{ge2021ota}   & \begin{tabular}[c]{@{}l@{}}Top-$r^2$ points closest  \\ to the object's center \end{tabular}                 & \begin{tabular}[c]{@{}l@{}}Cls., Loc. loss  \\ and center prior\end{tabular} & \begin{tabular}[c]{@{}l@{}}Modeling the scores as an optimal transportation cost. \\ Using Sinkhorn-Knopp Iteration to select top-k.\end{tabular}         \\ \hline
	\end{tabular}}
\end{table*}

Modern object detectors can be single or multi-stages. Single-stage is simpler and more efficient, while multi-stages are more complex but predict with higher precision. 

In \textbf{single stage detectors}, classification and localization are predicted concurrently for each pixel. There are anchor-based or anchor-free methods. For anchor-based detectors, such as SSD \cite{liu2016ssd}, RetinaNet \cite{lin2017focal}, an anchor is typically considered as positive if its IoU with the nearest ground truth box is at least 0.5, negative if less than 0.4, and ignored otherwise. ATSS \cite{Zhang2020} improves RetinaNet by selecting 9 nearest anchors from the object center in each feature level, and use the mean plus standard deviation (std) of IoU from this set as the assignment threshold. 
Anchor-free methods instead resort objects by points. FCOS \cite{tian2019fcos} assigns a point as positive if it is in a ground truth box, negative otherwise. If the point falls into two or more ground truth boxes, the smallest box is chosen. CornerNet \cite{law2018cornernet} represents an object by two corners of the bounding box, and only the corner points are assigned as positive. During training, the farther the points from the corner center, the larger weight it contributes to the loss function as negative samples.

In \textbf{multi-stage detectors}, the first stage network proposes candidate regions, by which the features are cropped and fed to the following stages for the box refinement and class prediction. The process can continue in a cascade manner as many stages as needed. Anchor assignments are also different in each stage. In the first stage, \ie proposal network, an anchor is typically considered as positive if its IoU with any ground-truth is greater than $0.7$, negative if less than $0.3$, and ignored otherwise. For the second stage, $0.5$ IoU threshold is used to separate positive and negative anchors. Cascade R-CNN \cite{cai2018cascade} improves Faster R-CNN by adding more stages. Setting the IoU threshold too high leads to extremely few positive samples, thus insufficient to train the network. Therefore, Cascade R-CNN re-assign the samples with IoU thresholds progressively increasing from $0.5$ to $0.7$ after each stage.


In summary, label assignments in the aforementioned methods are static and heuristically designed.

\subsection{Learning Label Assignment}
\label{sec:learninglabel}

Learning label assignment approaches update the network weights and learn the assignment iteratively towards final optimization. There are two main approaches.  

In \textbf{hard assignment}, samples can only be either positive or negative. The pipeline generally includes three steps. We first select and \textit{pre-assign} a set of potential samples as positive, then \textit{evaluate} their performance by a cost function. Finally, we select a threshold to separate the samples having lower cost as true positive, while the rest are \textit{reassigned} to negative or ignored.    \Tab{HardLearningCompare} compares different methods of learning label assignment in these aspects for single stage detector. For two-stage detectors, Dynamic-RCNN \cite{dynamicrcnn} progressively increases the IoU threshold in the second stage by mean of the predicted IoU of the proposal network, and modifies the hyper-parameter of SmoothL1 Loss function according to the statistics of regression loss.

In \textbf{soft assignment}, samples can have both positive and negative semantic characteristics. There are two main approaches: soft-weighted loss and soft-targets supervision. In the former, a set of positive candidates are first selected by a relaxed criterion, \eg low IoU threshold, then their importance scores are evaluated and used as weights to sum the loss values of the samples. FreeAnchor \cite{zhang2021learning} first constructs a bag of top-k anchors having highest IoU with object, then calculates the importance scores from the classification and localization confidence using Mean-Max function. Following this framework, NoisyAnchor \cite{li2020learning} derives the importance scores, \ie cleanliness, from the regression and classification losses of the samples. SAPD \cite{sapod} utilizes distance to object center as the metrics to measure the importance of samples. Furthermore, it leverages FSAF \cite{Zhu2019} module to assign soft-weights to samples in different feature levels. Alternatively, soft-target supervision decides how much a sample belonging to positive or negative based on an estimated quality distribution. AutoAssign \cite{autoassign} improves FCOS \cite{tian2019fcos} by estimating object's spatial distribution, \eg a 2D-Gaussian, for each category. Then, a point in the ground truth box is supervised by both positive and negative losses weighted by this distribution. Similarly, IQDet \cite{ma2021iqdet} builds a mixture of Gaussian models to capture IoU quality. Differently, positive points are randomly sampled by this distribution and via a bi-linear interpolation. The quality distribution is also used as the soft-label for classification supervision.

\subsection{Distillation in Object Detection}
\label{sec:distillation}
Distillation is first introduced in the seminal work of Hilton \etal \cite{hinton2015distilling} for image classification. By using the prediction of a teacher network as soft training targets, we can distill the knowledge from the teacher to a student. Distillation then becomes the principle for many other related problems, such as self-supervised learning. However, applying distillation to object detection is not straightforward, since classification and localization must be learned simultaneously. Zheng \etal \cite{zheng2021localization} hypothesize that a teacher is better in estimating the localization uncertainty, which can be used as the dark knowledge for distillation. However, the method is only applicable to Generalized Focal detector \cite{li2020generalized} due to its formulation problem. Other works focus on distillation by feature mimicking. Chen \etal \cite{chen2017learning} combine feature mimicking and soft-label distillation on Faster-RCNN. Other researches \cite{zagoruyko2016paying,wang2019distilling, li2017mimicking} believe that background features are noisy, thus propose applying semantic masks to focus attention on the meaningful regions, especially in and near foregrounds. In contrast, Guo \etal \cite{guo2021distilling} consider that background can capture the object's relationship and suggest decoupling the background features during distillation.

Although these methods exploit different ways for distillation, they all enforce a direct mimicking the teacher's final or intermediate outputs. This may be too restrictive if their architectures or performance are significantly different. Therefore, several papers \cite{mirzadeh2020improved, zheng2021localization} propose to add a teacher assistant network to bridge their gaps. Nevertheless, this introduces more hyperparameters, larger memory, longer training, and complicates the process.

Our proposed LAD extends the distillation concept and is orthogonal with these methods. Generally, they can be combined to further improve the performance.

\section{Method}
\subsection{Soft-label distillation} \label{sec:softlabel}
In network distillation, the teacher's prediction is referred as soft-label, which is believed to capture the network's dark knowledge and provide richer information for training a student.   
Adopting the convention from the classification problem \cite{hinton2015distilling}, we use the Kullback-Leibler (KL) divergence loss. Follow Focal loss \cite{lin2017focal}, we also add the focal term to deal with the extreme imbalance problem
\begin{equation}\label{eq:KLloss}
KL(p_t,p_s) = \sum\limits_{c=1}^{C} w^c \left(p_t^c \log\frac{p_t^c}{p_s^c} + (1-p_t^c)\log\frac{1-p_t^c}{1-p_s^c}\right),
\end{equation}
where $p_t$ and $p_s$ denote the teacher and student prediction (i.e. after Sigmoid), $c$ is the class index of total $C$ classes in the dataset, and $w^c\!=\!|p_t^c \!-\! p_s^c|^{\gamma}$ is the focal term. Note, when replacing the soft-label $p_t$ with an one-hot vector, \eqn{KLloss} becomes the standard Focal loss. Alternatively, we can also use $\mathcal{L}_1$ or $\mathcal{L}_2$ losses. Different from the classification losses, localization loss is only computed for positive samples. Hence, we select the predicted boxes having $IoU\!>\!0.5$ \wrt its nearest ground truth box for localization distilling. 

\subsection{Label Assignment Distillation}
\label{sec:lad}
LAD is very general and can be applied to many learning label assignment methods, such as those described in \sec{learninglabel}. For a concrete example, we adopt PAA to implement LAD. In PAA, for each object, we select a set of anchors $\{a_i\}$ that have $IoU\geq 0.1$, and evaluate the assignment cost   
\begin{equation}\label{eq:PAA_assign}
c_i = FL(p_i,C) + (1 - IoU(b_i,B))
\end{equation} 
as if they are positive anchors, where $FL$ is the Focal loss, $(p_i,b_i)$ are the predicted class probability and bounding box of $a_i$, $(C,B)$ are the ground truth class and bounding box of the object. Then, we build a distribution model for $\{c_i\}$ by a mixture of two Gaussian: $G(\mu_1,\sigma_1) + G(\mu_2,\sigma_2), \mu_1 < \mu_2$. We then assign anchors with $c_i <\mu_1$ as the true positive, and reassign the other as negative, as illustrated in \fig{PAA_diag}.

To convert PAA to LAD, we simply use the teacher's prediction $(p_T^i,b_T^i)$ instead of $(p_i,b_i)$ in \eqn{PAA_assign}, and proceed exactly the steps, as  illustrated in \fig{LAD_diag}. Consequently, LAD presents a new distillation method for object detection that works without directly mimicking the teacher's outputs.  

\subsection{A Pilot Study} \label{sec:PilotStudy} 
Our motivation is to seek a method that utilizes the knowledge from teacher network, and this can be achieved either by soft-label or LAD, as shown in \fig{Motivation}. Therefore, we conduct a pilot study to understand their properties.
\subsubsection{A preliminary comparision}\label{sec:Prelim} 
We first compare the two methods using PAA detector with ResNet backbone R101 and R50 as teacher and student, and report the results for the student PAA-R50 in \tab{Soft-vs-LAD}.
\begin{table}[!h]
	\caption{Compare the performance of the student PAA-R50 using Soft-Label, Label Assignment Distillation (LAD) and their combination (SoLAD) on COCO validation set. (*) denotes longer warming-up learning rate (3000 iterations). }
	\label{tab:Soft-vs-LAD}
	\center
	\small{
	\begin{tabular}{|l|c|c|c|}
		\hline
		Method          & $\gamma$   & mAP  & Improve \\ \hline
		Baseline PAA           & $2  $     & $40.4$ &   --      \\ \hline \hline
		Soft-Label - KL loss*  & $0  $     & $39.8$ & $-0.6$    \\ \hline
		Soft-Label - KL loss*  & $0.5$     & $41.3$ & $+0.9$    \\ \hline
		Soft-Label - KL loss   & $2.0$     & $39.6$ & $-0.8$    \\ \hline
		Soft-Label - $\mathcal{L}_1$ loss*  & --       & $40.4$ &  $0.0$    \\ \hline 
		Soft-Label - $\mathcal{L}_2$ loss   & --       & $41.0$ & $+0.6$    \\ \hline \hline
		\textbf{LAD} (ours)    & $2$       & $\textbf{41.6}$ & $\textbf{+1.2}$    \\ \hline \hline
		\textbf{SoLAD} (ours) - KL loss& $0.5$     & $\textbf{42.4}$ & $\textbf{+2.0}$    \\ \hline
	\end{tabular}}
\end{table}

\Tab{Soft-vs-LAD} shows that both methods can improve the baseline. For soft-label distillation, KL loss is better than $\mathcal{L}_1$ and $\mathcal{L}_2$ losses, but it must be tuned carefully. We found that a small positive $\gamma =0.5$ is critical to achieve good performance ($41.3\rm AP$). Notably, this makes the training unstable due to large initial error, which hence requires a much longer warming-up learning rate. In contrast, LAD yields the higher result ($41.6\rm AP$) with the standard setting. 

Finally, we show that LAD can be easily combined with soft-label distillation. Concretely, label assignment and training losses are conducted exactly as in LAD, but with additional soft-label distillation losses. We name this combination as \textbf{SoLAD}. As shown in \tab{Soft-vs-LAD}, SoLAD can improve the baseline PAA by $+2.0\rm AP$, which accumulates the improvement from each component: $+0.9\rm AP$ of soft label and $+1.2\rm AP$ of LAD. Remarkably, the student PAA-R50's performance ($42.4\rm AP$) closely converges to its teacher PAA-R101 ($42.6\rm AP$). 

The results above are very promising. It shows that LAD is simple but effective, orthogonal but complementary to other distillation techniques, and worth further exploration.

\subsubsection{Does LAD need a bigger teacher network?} 
Conventionally, a teacher is supposed to be the larger and better performance network. To verify this assumption, we conduct a similar experiment as in \sec{Prelim}, but swap the teacher and student. The results are shown in \tab{SmallTeacher}.

\begin{table}[!h]
	\caption{Compare Soft-Label and Label Assignment Distillation (LAD) on COCO validation set. Teacher and student use ResNet50 and ResNet101 backbone, respectively. \2x denotes the \2x training schedule. }
	\label{tab:SmallTeacher}
	\small{
	\begin{tabular}{|l|c|c|c|c|}
		\hline
		\textbf{Method }       & \textbf{Teacher} & \textbf{Network}  & \textbf{mAP}  & \textbf{Improve} \\ \hline
		Baseline & None    & PAA-R101 & $42.6$ & --       \\ \hline
		Baseline (\2x) & None    & PAA-R101 & $43.5$ & $+0.9$    \\ \hline \hline 
		Soft-Label    & PAA-R50   & PAA-R101 & $40.4$  & $-2.2$    \\ \hline
		\textbf{LAD (ours)}     & PAA-R50 & PAA-R101 & $\textbf{43.3}$ & $\textbf{+0.7}$  \\ \hline
	\end{tabular}}
\end{table}

\Tab{SmallTeacher} demonstrates a distinctive advantage of LAD. Specifically, using the teacher PAA-R50 ($40.4\rm AP$) with soft-label distillation, the student PAA-R101's performance drops by $-2.2\rm AP$ relative to the baseline. In contrast, LAD improves the student by $+0.7\rm AP$ and achieves $43.3\rm AP$, asymptotically reaching the baseline trained with \2x schedule ($43.5\rm AP$). Especially, the student surpasses its teacher with a large margin ($+2.9\rm AP$). This proves a very useful and unique property of LAD - a bidirectional enhancement, because potentially any teacher now can improve its student regardless of their performance's gap. 


\subsection{Co-learning Label Assignment Distillation} 
\begin{figure*}[!t]
	\centering
	\includegraphics[width=0.95\linewidth]{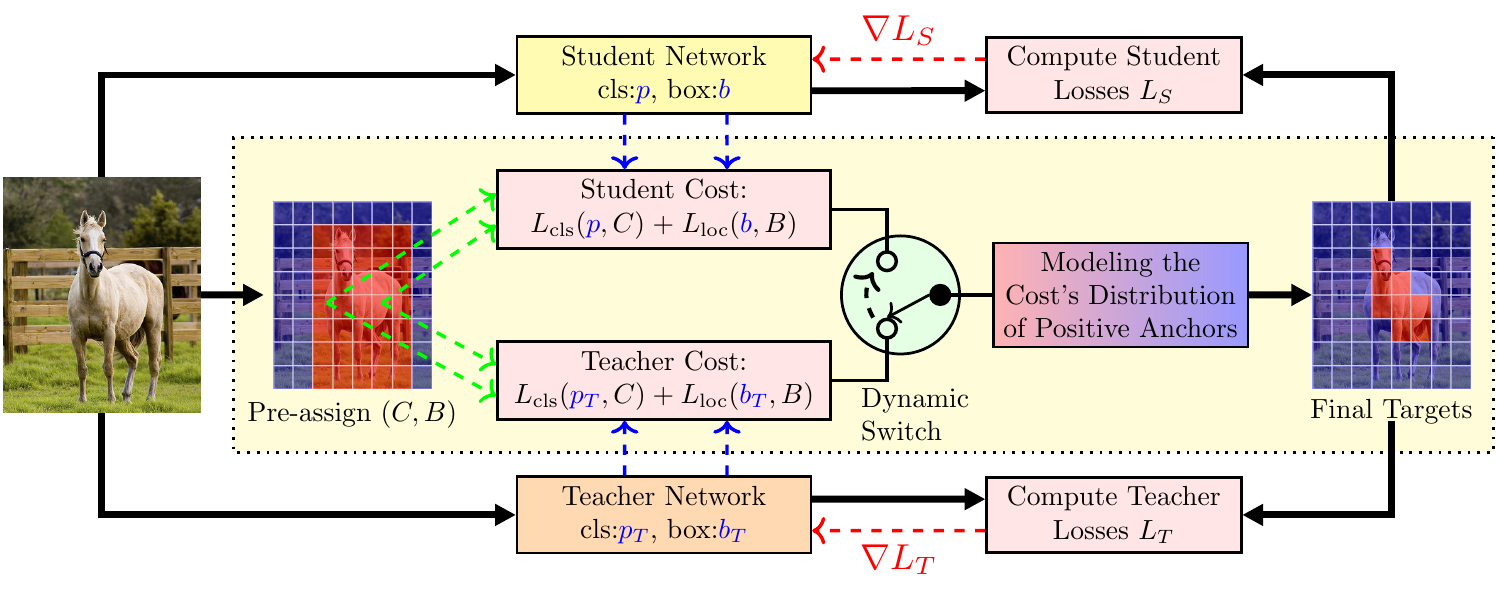}
	\caption{Co-Learning Label Assigment Distillation (CoLAD) framework. Dynamic Switch is described in \eqn{Switch}.}
	\label{fig:CoLAD}
\end{figure*} 
The pilot study in \sec{PilotStudy} proves that the proposed LAD is a simple and effective solution to improve network training for object detection. However, like many other distillation methods, it requires a teacher pretrained in advance. 
Learning label assignment methods \cite{Carion2020, kim2020probabilistic, autoassign, ge2021ota} on the other hand, use a bootstrap mechanism to self-distill the labels. However, a potential drawback is that it can be trapped in local minima. This is because it only exploits the highest short-term reward, i.e. lowest assignment cost, without any chance for exploration.

Therefore, to avoid these drawbacks and combine their advantages, we propose the Co-learning Label Assignment Distillation (CoLAD). CoLAD does not need a beforehand teacher, while potentially avoiding falling into local minima. The framework is illustrated in \fig{CoLAD}. Concretely, we use two separate networks similarly to LAD, but none of them must be pre-trained. Instead, we train them concurrently, and dynamically switch the role of teacher and student based on their performance indicator $\rho$.  We propose two criteria for $\rho$, namely Std/Mean score ($\rho _{\sigma/\mu}$) and Fisher score ($\rho_{\rm Fisher}$).
Let $\{c_i\}$ be the set of the assignment costs  evaluated by a network on the pre-assigned positive anchors, the criteria are respectively defined as:
\begin{equation}\label{eq:Switch}
\rho _{\sigma/\mu} = \frac{\sigma}{\mu}, \hspace{5mm} \rho_{\rm Fisher}= \frac{(\mu^+ - \mu^-)^2}{{\sigma^+}^2 + {\sigma^-}^2}, 
\end{equation}
where $(\mu,\sigma)$ are the mean and std. of $\{c_i\}$, and in the Fisher Score, we approximate the distribution of $\{c_i\}$ by a mixture of two Gaussian models $G(\mu^+,\sigma^+) +  G(\mu^-,\sigma^-)$.  

At each iteration, the network with higher $\rho$ can be selected as the teacher. Therefore, by dynamically switching the teacher-student, the process no longer relies on a single network, stochastically breaking the loop of training-self assignment. This also brings more randomness at the initial training due to network initialization, hence potentially reducing the risk of local minimal trapping. 

Finally, we can also apply the dynamic switch \eqn{Switch} to LAD, in which the teacher is pretrained and fixed. This allows the student learn from the teacher at the early training stage. However, when the student is gradually improved and potentially surpasses its teacher, it can switch to the self-distillation mode. This makes LAD more adaptive.


\section{Experiments}
We evaluate the proposed methods on the MS COCO benchmark \cite{COCO}, which has about 118k images in the \textit{train} set, 5k in the \textit{val} set and 20k in the \textit{test-dev} set. Following the common protocol, we report the ablation studies on the \textit{val} set, and compare with previous methods on the \textit{test-dev} set using the official COCO evaluation tool.

\subsection{Implementation Details}
We use PAA detector with different ResNet backbones R18, R50, R101 \cite{he2016deep} with FPN \cite{lin2017feature} for the baseline. The backbones are initialized with weights pre-trained on ImageNet \cite{deng2009imagenet}. Our code is implemented using MMDetection framework \cite{chen2019mmdetection}, and follows its standard setup. Specifically, we use SGD optimizer with weight decay of 0.0001 and the momentum of 0.9. $1 \times $ training schedule includes 12 epochs with initial learning rate of 0.01, which is then divided by 10 at epoch 8 and epoch 11. $2 \times $ training schedule includes 24 epochs with initial learning rate of 0.01, which is then divided by 10 at epoch 16 and epoch 22. Random horizontal flipping is used in data augmentation. We use batch size 16, with image size of $(800 \times 1333)$ pixels for training and testing. All other hyperparameters, such as loss weights and post-processing, are identical to the PAA baseline.

\subsection{Ablation Study for CoLAD}
In CoLAD, the roles of teacher-student are not persistent, so the terminology teacher and student networks may not exactly apply. However, for convenience, we still call the network with higher performance at initial as the teacher. If two networks are not pretrained, then the one having more parameters and layers is supposed to be the initial teacher.
\subsubsection{Switching Criteria in CoLAD}
First, we want to know which switching criteria, the Std/Mean or Fisher score, is more suitable for CoLAD, and how CoLAD performs in comparison with the baseline PAA and LAD. Since LAD uses teacher PAA-R50 pretrained with \1x schedule, we also adopt it in this experiment. We train the student PAA-R101 for \1x and \2x schedule.

\begin{table}[!h]
	\caption{Compare criteria to dynamically switch teacher and student. PAA-R50 pre-trained by \1x schedule on COCO is used as the initial teacher in LAD and CoLAD.}
	\label{tab:dlad-criteria}
	\centering
	\small{
	\begin{tabular}{|l|l|l|c|}
		\hline
		\textbf{Student }  & \textbf{Method}          & \textbf{mAP} & \textbf{Improve} \\ \hline
		\multirow{4}{*}{\begin{tabular}[c]{@{}l@{}}PAA-R101\\ (train \1x)\end{tabular}} 
		& PAA Baseline     & $42.6$ & --\\ \cline{2-4} 
		& LAD              & $43.3$ & $+0.7$\\ \cline{2-4} 
		& CoLAD - Std/Mean & $43.3$ & $+0.7$ \\ \cline{2-4} 
		& CoLAD - Fisher   & $\textbf{43.4}$ & $\textbf{+0.8}$ \\ \hline
		\multirow{2}{*}{\begin{tabular}[c]{@{}l@{}}PAA-R101\\ (train \2x)\end{tabular}} 
		& PAA Baseline     & $43.5$ & -- \\ \cline{2-4} 
		& CoLAD - Std/Mean & $43.9$ & $+0.4$ \\ \cline{2-4} 
		& CoLAD - Fisher   & $43.9$ & $+0.4$ \\ \hline
	\end{tabular}}
\end{table}

As reported in \tab{dlad-criteria}, the criteria yield similar results in both \1x and \2x schedule training. Hence, we use the Std/Mean as the default criterion in all later experiments, since it is simpler and less computationally expensive than the Fisher score. For \1x schedule, LAD and CoLAD perform equivalent and improve the baseline by $+0.7\rm AP$.  

\subsubsection{Impact of network's pretrained weights}
Assuming no pretrained teacher, we ask whether it is better to train two networks with CoLAD, rather than conventionally training each one separately.  
We also try to find how much they are improved, if one of them is already pretrained in COCO. To answer these questions, we train the pair PAA-R50 and PAA-R101 for \1x and \2x schedule with different initialization, and report the results in \tab{CoLADr50-r101}. 

\begin{table}[!h]
	\caption{CoLAD training of PAA-R50 and PAA-R101 detectors, with different pretrained weights.  \1x/\2x denotes that the models are trained or pretrained on COCO with \1x/\2x schedule, respectively. (*) denotes comparing with the Baseline trained with \2x schedule.}
	\label{tab:CoLADr50-r101}
	\centering
	\begin{adjustbox}{max width=\columnwidth}
	\small{
	\begin{tabular}{|l|l|l|l|c|}
		\hline
		\textbf{Method }                                               & \textbf{Networks}                                                    & \textbf{Init.}                                      & \textbf{mAP}                                                 & \textbf{Improve}                                                  \\ \hline
		Baseline                                              & PAA-R50                                                    & \1x/\2x                                           & $40.4/ 41.6$                                          & --                                                        \\ \hline
		Baseline                                              & PAA-R101                                                   & \1x/\2x                                           & $42.6/ 43.5$                                          & --                                                        \\ \hline
		\begin{tabular}[c]{@{}l@{}}CoLAD \\ (\1x)\end{tabular}& \begin{tabular}[c]{@{}l@{}}PAA-R50\\ PAA-R101\end{tabular} & \begin{tabular}[c]{@{}l@{}}No\\ No\end{tabular} & \begin{tabular}[c]{@{}l@{}}$41.3$\\ $43.1$\end{tabular} & \begin{tabular}[c]{@{}l@{}}$+0.9$\\ $+0.5$\end{tabular}      \\ \hline
		\begin{tabular}[c]{@{}l@{}}CoLAD \\ (\1x)\end{tabular} & \begin{tabular}[c]{@{}l@{}}PAA-R50\\ PAA-R101\end{tabular} & \begin{tabular}[c]{@{}l@{}}No\\ \1x\end{tabular} & \begin{tabular}[c]{@{}l@{}}$41.6$\\ $44.1$\end{tabular} & \begin{tabular}[c]{@{}l@{}}$+1.2$\\ $+0.6$*\end{tabular}     \\ \hline
		\begin{tabular}[c]{@{}l@{}}CoLAD \\  (\1x)\end{tabular}  & \begin{tabular}[c]{@{}l@{}}PAA-R50\\ PAA-R101\end{tabular} & \begin{tabular}[c]{@{}l@{}}\1x\\ No\end{tabular} & \begin{tabular}[c]{@{}l@{}}$42.6$\\ $43.3$\end{tabular} & \begin{tabular}[c]{@{}l@{}}$+1.0$ *\\ $+ 0.7$\end{tabular}   \\ \hline
		\begin{tabular}[c]{@{}l@{}}CoLAD \\ (\2x)\end{tabular} & \begin{tabular}[c]{@{}l@{}}PAA-R50\\ PAA-R101\end{tabular} & \begin{tabular}[c]{@{}l@{}}No\\ No\end{tabular} & \begin{tabular}[c]{@{}l@{}}$42.4$\\ $44.1$\end{tabular} & \begin{tabular}[c]{@{}l@{}}$+0.8$ *\\ $+ 0.6$ *\end{tabular} \\ \hline
	\end{tabular}}
	\end{adjustbox}
\end{table}

As shown in \tab{CoLADr50-r101}, when no pretrained teacher available (\nth{3} and \nth{6} row), \textit{networks trained with CoLAD outperfom those that were trained independently}. Specifically, there are $+0.9/+0.5 \rm AP$ improvement in \1x schedule (\nth{3} \vs \nth{1} row), and $+0.8/+0.6\rm AP$ in \2x schedule (\nth{6} \vs \nth{2} row) compared to the baseline PAA-R50/R101, respectively. Although the improvement in PAA-R50 may be expected thanks to co-training with a larger PAA-R101, the fact that the network PAA-R101 also got improved thanks to its smaller partner PAA-R50 is remarkable. 

Secondly, when one of the networks is pretrained on COCO (\nth{4} and \nth{5} rows), the improvements are better than without pretraining (\nth{3} row), which is expected. However, the improvements are somewhat marginal for the students. In \1x training, CoLAD improves the students PAA-R50 from $41.3$ to $41.6\rm AP$ ($+0.3$) (\nth{3} \vs \nth{4} row), and PAA-R101 from $43.1$ to $43.3\rm AP$($+0.2$) (\nth{3} \vs \nth{5} row). In \2x training, the improvements are from $42.4$ to $42.6\rm AP$($+0.2$) for R50 (\nth{6} \vs \nth{5} row), while R101 reaches to $44.1\rm AP$ (\nth{6} and \nth{4} row). This demonstrates that \textit{CoLAD can replace LAD when a pretrained teacher is not available.} 

What's more interesting is that \textit{``one plus one is better than two"}, \ie a teacher which was trained for \1x schedule, after joining CoLAD for another \1x, outperforms the one independently trained for \2x schedule. Concretely, CoLAD improves the teachers PAA-R50 from $41.6$ to $42.6\rm AP$($+1.0$) (\nth{1} \vs \nth{5} row), and PAA-R101 from $43.5$ to $44.1\rm AP$($+0.6$) (\nth{2} \vs \nth{4} row). 
Finally, the experiments above support our hypothesis that the dynamic switching mechanism of CoLAD reduces the risk of local minima trapping, and that is why it outperforms the baseline PAA.

\subsubsection{Impact of Teacher-Student's relative model sizes}
To understand how the relative gap between the model sizes can affect the student's performance, we compare the results of two model pairs, including (PAA-R18, PAA-R50) and (PAA-R18, PAA-R101) using LAD and CoLAD. 

\begin{table}[!h]
	\caption{Evaluate the teacher-student's relative model sizes. The student PAA-R18 is trained with different teachers. \1x means teacher is pretrained on COCO with \1x schedule.}
	\label{tab:relative-size}
	\centering
	\small{
		\begin{tabular}{|c|c|c|c|c|}
			\hline
			\textbf{Teacher}          & \textbf{Init.} & \textbf{Method} & \textbf{mAP} & \textbf{Improve} \\ \hline
            None                      & --              & Baseline PAA    & $35.8$         & --                \\ \hline
			\multirow{2}{*}{PAA-R50}  & \1x             & LAD             & $36.9$         & $+1.1$             \\ \cline{2-5} 
			& No                                        & CoLAD           & $36.5$         & $+0.7$             \\ \hline
			\multirow{2}{*}{PAA-R101} & \1x             & LAD             & $36.8$         & $+1.0$             \\ \cline{2-5} 
			& No                                        & CoLAD           & $36.6$         & $+0.8$             \\ \hline
		\end{tabular}
	}
\end{table}
 
\Tab{relative-size} reports the experiments. We see that the impact of teachers with different capacity and pretrained weights is negligible and in the typical noise $\pm0.1 \rm AP$. This contradicts with the paradox \cite{mirzadeh2020improved} that a student's performance degrades when the gap with its teacher is large. This proves that the discrepancy between teacher and student may not be a weakness of LAD and CoLAD. However, from a different viewpoint, this is also a drawback because a better teacher can't improve the student further. Hence, we should also combine LAD with other distillation techniques in order to take the full advantages of the teacher.    

\subsection{Comparison with State-of-the-art}
\label{sec:sota}
We compare our proposed CoLAD with other methods on MS COCO \textit{test-dev} set, especially the recent ones addressing the label assignment for single-stage detectors. Following the previous works \cite{kim2020probabilistic, ge2021ota, autoassign}, we train the model with \2x scheduler, randomly scale the shorter size of the image into the range of 640 and 800. We use the PAA-R50 trained with $3\times$ schedule and multi-scale on COCO as the initial teacher \footnote{PAA-R50's pretrained weight is provided by MMDetection  https://github.com/open-mmlab/mmdetection/tree/master/configs/paa}. The teacher was evaluated with $43.3\rm AP$ and $43.8\rm AP$ on the \textit{minval} and \textit{test-dev} sets, respectively. 

\begin{table}[!h]
\caption{Compare with state-of-the-art (SOTA) methods (single model trained \2x schedule) on COCO \textit{test-dev} set.}
\label{tab:SOTA}
\centering
\begin{adjustbox}{max width=\columnwidth}
\small{
\begin{tabular}{|l|l|l|l|l|l|l|}
    \hline
    \multicolumn{1}{|l|}{\textbf{Method}} & \multicolumn{1}{l|}{\textbf{AP}} & \multicolumn{1}{l|}{\textbf{AP$_{\textbf{50}}$}} & \multicolumn{1}{l|}{\textbf{AP$_{\textbf{75}}$}} & \multicolumn{1}{l|}{\textbf{AP$_\textbf{S}$}} & \multicolumn{1}{l|}{\textbf{AP$_\textbf{M}$}} & \multicolumn{1}{l|}{\textbf{AP$_\textbf{L}$}} \\ \hline
    \multicolumn{7}{|c|}{\textbf{ResNet-101 backbone}} \\
    FCOS\cite{tian2019fcos}			& 41.5          & 60.7          & 45.0          & 24.4          & 44.8          & 51.6 \\
    N.Anchor\cite{li2020learning}	& 41.8          & 61.1          & 44.9          & 23.4          & 44.9          & 52.9 \\
    F.Anchor\cite{zhang2019freeanchor}		& 43.1          & 62.2          & 46.4          & 24.5          & 46.1          & 54.8 \\
    SAPD \cite{sapod}				& 43.5          & 63.6          & 46.5          & 24.9          & 46.8          & 54.6 \\
    MAL\cite{Ke2020}				& 43.6          & 61.8          & 47.1          & 25.0          & 46.9          & 55.8 \\
    ATSS\cite{Zhang2020}			& 43.6          & 62.1          & 47.4          & 26.1          & 47.0          & 53.6 \\
    GFL\cite{li2020generalized}		& 45.0          & 63.7          & 48.9          & 27.2          & 48.8          & 54.5 \\
    A.Assign\cite{autoassign}		& 44.5          & 64.3          & 48.4          & 25.9          & 47.4          & 55.0 \\
    PAA\cite{kim2020probabilistic}	& 44.8          & 63.3          & 48.7          & 26.5          & 48.8          & 56.3 \\
    OTA\cite{ge2021ota}				& 45.3          & 63.5          & 49.3          & 26.9          & 48.8          & 56.1 \\
    IQDet\cite{ma2021iqdet}			& 45.1          & 63.4          & 49.3          & 26.7          & 48.5          & 56.6 \\
    \textbf{CoLAD[ours]}            & \textbf{46.0} & \textbf{64.4} & \textbf{50.6} & \textbf{27.9} & \textbf{49.9} & \textbf{57.3} \\ \hline
    \multicolumn{7}{|c|}{\textbf{ResNeXt-64x4d-101 backbone}} \\
    FSAF \cite{Zhu2019}             & 42.9			& 63.8			& 46.3			& 26.6			& 46.2			& 52.7	\\
    FCOS\cite{tian2019fcos}			& 43.2			& 62.8			& 46.6			& 26.5			& 46.2			& 53.3	\\
    F.Anchor\cite{zhang2019freeanchor}		& 44.9			& 64.3			& 48.5			& 26.8			& 48.3			& 55.9	\\
    SAPD\cite{sapod}				& 45.4			& 65.6			& 48.9			& 27.3			& 48.7			& 56.8	\\
    ATSS\cite{Zhang2020}			& 45.6			& 64.6			& 49.7			& 28.5			& 48.9			& 55.6	\\
    A.Assign\cite{autoassign}		& 46.5			& 66.5			& 50.7			& 28.3			& 49.7			& 56.6	\\
    PAA\cite{kim2020probabilistic}	& 46.6			& 65.6			& 50.8			& 28.8			& 50.4			& 57.9	\\
    OTA\cite{ge2021ota}				& 47.0			& 65.8			& 51.1			& 29.2			& 50.4			& 57.9	\\
    IQDet\cite{ma2021iqdet}			& 47.0			& 65.7			& 51.1			& 29.1			& 50.5			& 57.9	\\
    \textbf{CoLAD[ours]}            & \textbf{47.5} & \textbf{66.4} & \textbf{52.1} & \textbf{29.8} & \textbf{51.0} & \textbf{59.1} \\ \hline
    \multicolumn{7}{|c|}{\textbf{ResNeXt-64x4d-101-DCN backbone}} \\
    SAPD\cite{sapod}				& 47.4			& 67.4			& 51.1			 & 28.1			 & 50.3          & 61.5          \\
    ATSS\cite{Zhang2020}			& 47.7			& 66.5			& 51.9			 & 29.7			 & 50.8          & 59.4          \\
    A.Assign\cite{autoassign}		& 48.3			& 67.4			& 52.7			 & 29.2			 & 51.0          & 60.3          \\
    PAA\cite{kim2020probabilistic}	& 49.0			& 67.8			& 53.3			 & 30.2			 & 51.3          & 62.2          \\
    OTA\cite{ge2021ota}				& \textbf{49.2} & 67.6			& 53.5			 & 30.3			 & 52.5 & \textbf{62.3} \\
    IQDet\cite{ma2021iqdet}			& 49.0			& 67.5			& 53.1			 & 30.0			 & 52.3          & 62.0          \\
    \textbf{CoLAD[ours]}			& \textbf{49.2}			& \textbf{68.3}	& \textbf{54.1} & \textbf{30.6} & \textbf{52.8} & 61.9 		 \\ \hline
\end{tabular}}
\end{adjustbox}
\end{table}

Inspired by \cite{autoassign, chen2021you, zhou2021probabilistic}, we also modify the PAA's head architecture by connecting the classification and localization branches by an auxiliary ``objectness" prediction, which is supervised implicitly.  However, we use a slightly different implement with \cite{autoassign, chen2021you, zhou2021probabilistic}, which we call Conditional Objectness Prediction (COP). Since COP is not our main contribution, we refer the reader to Appendix A for the motivation and more ablation studies. We report the results of the PAA network amended with COP and trained with CoLAD in \tab{SOTA}.

With R101 backbone, CoLAD achieves $46.0 \rm AP$,  consistently outperforming other recent methods for label assignment, such as OTA \cite{ge2021ota} ($45.3  \rm AP$), IQDet \cite{ma2021iqdet} ($45.1  \rm AP$) and PAA \cite{kim2020probabilistic} ($44.8 \rm AP$) by large margins on all evaluation metrics. With the larger backbone ResNeXt-64x4d-101, our model can further improve to $47.5 \rm AP$ and surpasses all  existing methods. Finally, for the ResNeXt-64x4d-101 with Deformable Convolutional Networks (DCN) \cite{zhu2018deformable}, CoLAD and OTA both achieve the highest score $49.2 \rm AP$. Note that, with this DCN backbone, the top 4 methods perform very similar ($\pm 0.2\rm AP$). 

Nevertheless, it is not our intention for LAD/CoLAD to become a replacement for current SOTA methods. Instead, we find it useful to use a simple technique, with a smaller and lower performance teacher ResNet50, to successfully boost the current SOTA results. Indeed, as shown in the next section, by simply replacing ResNet50 with a stronger teacher, we can still improve the results above further.

\subsection{Vision Transformer Backbone}
\label{sec:swin-transformer-backbone}
Vision Transformer recently demonstrates its potential to replace convolution networks. Therefore, we conduct experiments to verify the generalization of LAD with Transformer backbones. To this end, we select the Swin-B Transformer \cite{liu2021Swin} and use the teacher PAA-ResNeXt-64x4-101-DCN obtained in the section \ref{sec:sota}. Since CoLAD training with these two networks is extremely heavy, we use LAD but with the dynamic switch instead. We follow exactly the training setting in \cite{liu2021Swin}, such as Swin-B backbone is pre-trained on ImageNet, multi-scale training, batch-size 16, AdamW optimizer, and $3\times$ schedule. 

\begin{table}[!h]
	\caption{Compare the teacher PAA-ResNeXt101-DCN and the student PAA-Swin-B on COCO \textit{test} set.}
	\label{tab:swin_transformer_compare2}
	\centering
	\begin{adjustbox}{max width=\columnwidth}
		\small{
			\begin{tabular}{|l|l|l|l|l|l|l|}
				\hline
				\multicolumn{1}{|l|}{\textbf{Method}} & \multicolumn{1}{l|}{\textbf{AP}} & \multicolumn{1}{l|}{\textbf{AP$_{\textbf{50}}$}} & \multicolumn{1}{l|}{\textbf{AP$_{\textbf{75}}$}} & \multicolumn{1}{l|}{\textbf{AP$_\textbf{S}$}} & \multicolumn{1}{l|}{\textbf{AP$_\textbf{M}$}} & \multicolumn{1}{l|}{\textbf{AP$_\textbf{L}$}} \\ \hline
				Teacher -DCN	 & $49.2$			& $68.3$	& $54.1$ & $30.6$ & $52.8$ & $61.9$ \\ \hline
				\textbf{Student-Swin}  & $\textbf{52.0}$      & $\textbf{71.3}$      & $\textbf{57.2}$      & $\textbf{33.8}$      & $\textbf{55.7}$      & $\textbf{65.1}$ \\ \hline
		\end{tabular}}
	\end{adjustbox}
\end{table}
As reported in \tab{swin_transformer_compare2}, the student PAA-Swin-B once again surpasses its teacher with a large margin ($+2.8\rm AP$). Furthermore, on COCO \textit{val} set, it achieves $51.4\rm AP$, approaching the heavy Cascade Mask-RCNN ($51.9\rm AP$) \cite{liu2021Swin} using the same backbone, but without training segmentation. Finally, we recycle the PAA-Swin-B as a teacher to train PAA-R50/R101 with SoLAD, and compare the results to other distillation methods in \tab{swin_transformer_teacher}. 

Note that, it is not advisable to judge which method is better based on \tab{swin_transformer_teacher}, since each serves for a particular detector, thus having different baselines. Nevertheless, our SoLAD still achieves the highest performance for the same backbone and training schedule. 

\begin{table}[!h]
	\caption{Using PAA-Swin-B Transformer teacher to train PAA-R50/R101 with SoLAD. $\dagger$ denotes PAA head amended with COP. Results are compared with other methods using the same backbone on COCO \textit{test} set.}
	\label{tab:swin_transformer_teacher}
	\centering
	\begin{adjustbox}{max width=\columnwidth}
		\small{
			\begin{tabular}{|l|l|l|l|l|l|l|}
				\hline
				\multicolumn{1}{|l|}{\textbf{Method}} & \multicolumn{1}{l|}{\textbf{AP}} & \multicolumn{1}{l|}{\textbf{AP$_{\textbf{50}}$}} & \multicolumn{1}{l|}{\textbf{AP$_{\textbf{75}}$}} & \multicolumn{1}{l|}{\textbf{AP$_\textbf{S}$}} & \multicolumn{1}{l|}{\textbf{AP$_\textbf{M}$}} & \multicolumn{1}{l|}{\textbf{AP$_\textbf{L}$}} \\ \hline
				\multicolumn{7}{|c|}{\textbf{ResNet50 backbone - train \1x schedule}} \\  
				\textbf{SoLAD}            			& 42.9 & 62.2 & 46.8 & 25.1 & 46.5 & 54.0 \\
				\textbf{SoLAD} $\dagger$  			& \textbf{43.7} & \textbf{63.0} & \textbf{48.0} & \textbf{25.8} & \textbf{47.5} & \textbf{54.9} \\  \hline    
				FGFI \cite{wang2019distilling} 		& 39.9 & --   & --   & 22.9 & 43.6 & 52.8 \\
				TADF \cite{sun2020distilling}   	& 40.1 & --   & --   & 23.0 & 43.6 & 53.0 \\
				DeFeat \cite{guo2021distilling} 	& 40.9 & --   & --   & 23.6 & 44.8 & 53.0 \\ 	
				LD-GF \cite{zheng2021localization} 	& 41.2 & 58.8 & 44.7 & 23.3 & 44.4 & 51.1 \\ 
				GID \cite{dai2021general} 			& 42.0 & 60.4 & 45.5 & 25.6 & 45.8 & 54.2 \\ 
				\hline	\hline	
				\multicolumn{7}{|c|}{\textbf{ResNet101 backbone - train \2x schedule}} \\ 
				\textbf{SoLAD} $\dagger$ 					& \textbf{47.9} & \textbf{67.1} & \textbf{52.7} & \textbf{29.6} & \textbf{51.9} & \textbf{59.6} \\ \hline	
				GF2 \cite{li2021generalized} 		& 46.2 & 64.3 & 50.5 & 27.8 & 49.9 & 57.0 \\	
				LD-GF2 \cite{zheng2021localization} & 46.8 & 64.5 & 51.1 & 28.2 & 50.7 & 57.8 \\ 
				\hline
		\end{tabular}}
	\end{adjustbox}
\end{table}



\section{Conclusion}
This paper introduced a general concept of label assignment distillation (LAD) for object detection, in which we use a teacher network to assign training labels for a student. Different from all previous distillation methods which force the student to learn directly from the teacher's output, LAD indirectly distills the teacher's experience through the cost of the label assignment, thus defining more accurate training targets. We demonstrate a number of advantages of LAD, notably that it is very simple and effective, flexible to use with most of detectors, and complementary to other distillation techniques. Later, we introduced the Co-learning dynamic Label Assignment Distillation (CoLAD) to allow two networks to be trained mutually based on a dynamic switching criterion. Experiments on the challenging MS COCO dataset show that our method significantly surpasses all the recent label assignment methods. We hope that our work on LAD provides a new insight when developing object detectors, and believe that it is a topic worth further exploration.


\section*{Acknowledgement} 
We would like to thank Francisco Menendez, Su Huynh, Vinh Nguyen, Nam Nguyen and other colleagues for valuable discussion and helping reviewing the paper.
{\small
\bibliographystyle{ieee_fullname}
\bibliography{LADbib}
}
\begin{appendices}
\section{Conditional Objectness Prediction}
\subsection{Motivation and Network Architecture}
Since object detection performs classification and localization concurrently, their quality must be consistent. For example, a prediction with a high classification probability but low IoU box yields a false positive, while the reverse induces false negative. However, single-stage detectors implement the branches independently, typically each with 4 stacked convolutions. During training and inference, there is no connection between them. 

Furthermore, although the two branches have the same computation and feature extraction capacity, the localization receives significantly less training feedback than the classification. This is because most of the samples are negative, which have no box targets for training localization, hence are discarded during gradient backward. 
Recent methods also add auxiliary branches to predict the localization quality, such as IoU \cite{wu2020iou, kim2020probabilistic}, Centerness \cite{li2021generalized, zhang2019freeanchor, tian2019fcos}, but is trained only on positive samples. In addition, the positive samples of the same object are often connected and appear in a small local window. However, they are treated independently during non-maxima suppression.

\begin{figure}[!t]
	\center
	\includegraphics[width=\linewidth]{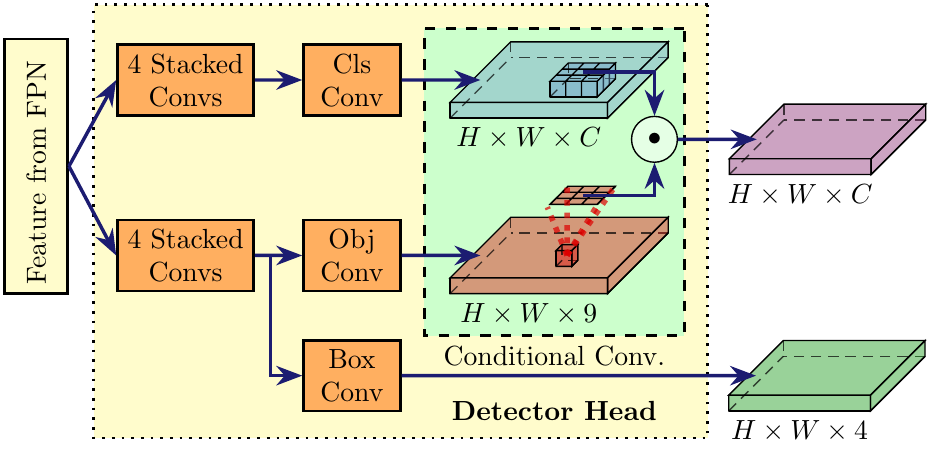}
	\caption{PAA head amended with Conditional Objectness Prediction (COP)}
	\label{fig:COP_diag}
\end{figure}

Therefore, we propose adding an auxiliary Conditional Objectness Prediction (COP) to the localization branch. It is similar to the Regional Proposal Network (RPN) of two-stage detector \cite{ren2015faster} but with renovations, as shown in \fig{COP_diag}. 
Concretely, at each anchor $a_i$, we predict the objectness scores $\{o_i^k\}_{k=1}^{3\times3}$ of its $3\times3$ nearest neighbors to capture their spatial correlation. The final classification probability is the dot product of the objectness $\{o_i^k\}$ and the corresponding $3\times3$ local window of the classification prediction $\{p_i^k\} _{k=1}^{3\times3}$
\begin{equation}
p(a_i) = \frac{1}{9}\sum_{k=1}^{3\times3} o_i^k p_i^k,
\end{equation}
where $o_i$ and $p_i$ are the confidence score (\ie after Sigmoid) of the objectness branch and classification branch, which are supervised implicitly and mutually through COP product during gradient back-propagation. Therefore, we can fuse and jointly train the branches, make the training consistent with inference. Consequently, all samples in the localization now receive gradient feedback. 

Our COP shares a common with the Implicit Object recently introduced in \cite{autoassign, chen2021you}, as they are both trained jointly with the classification branch. However, our motivation and implementation are different: (i) We believe features in the regression branch are also helpful to predict objects in the class-agnostic manner, similar to the RPN head in Faster-RCNN, and should not be discarded. COP is introduced to distribute gradient feedback to all samples in the localization branch. (ii) We implement COP as Conditional Convolution \cite{yang2019condconv, tian2020conditional}, where the weights are generated dynamically for each sample. Hence, we can embed the local relationship between the samples to reduce false-positive prediction.

\subsection{Ablation Study}
We investigate the effectiveness of the Conditional Objectness (COP) with different backbones, including EfficientNet-B0 (Eff-B0) \cite{tan2019efficientnet}, RepVGG-A0 (A0)\cite{ding2021repvgg}, ResNet18 (R18), and ResNet50 (R50)\cite{he2016deep}, and compare it with IoU prediction and Implicit Object  prediction (IOP). For easy comparison, we use the baseline PAA method, that has IoU prediction by default. 
\Tab{cop} summarizes the results.  
 
\textbf{\begin{table}[!ht]
	\centering
	\caption{Compare different auxiliary predictions: IoU, Implicit Object Prediction (IOP), and Conditional Objectness Prediction(COP) with different backbones. (*) denotes the branch is trained but not used during inference.}
	\label{tab:cop}
\begin{tabular}{|c|c|c|c|c|c|c|}
	\hline
	\multicolumn{3}{|c|}{\textbf{Auxiliary Prediction}} & \multicolumn{4}{c|}{\textbf{mAP}}                             \\ \hline 
	\textbf{IOU} & \textbf{\textbf{IOP}} & \textbf{COP} & \textbf{Eff-B0} & \textbf{A0}   & \textbf{R18}  & \textbf{R50}  \\ \hline 
	\checkmark      &                &                & 32.4          & 34.0          & 35.8          & 40.4          \\ \hline 
	\checkmark      & \checkmark     &                & 33.4          & 34.7          & 36.7          & \textbf{41.6} \\ \hline 
	\checkmark  &                       & \checkmark   & \textbf{33.5}   & \textbf{34.8} & \textbf{36.9} & \textbf{41.6} \\ \hline  \hline
	 *       & \checkmark     &                & 33.4          & \textbf{34.8} & 36.7          & 41.5          \\ \hline 
	 *      &                & \checkmark     & 33.5          & \textbf{34.8} & \textbf{36.9} & \textbf{41.6} \\ \hline \hline
	& \checkmark     &                & 33.3          & \textbf{34.8} & 36.6          & 41.1          \\ \hline 
	&                & \checkmark     & \textbf{33.4} & 34.7          & \textbf{36.9} & \textbf{41.2} \\ \hline 
\end{tabular}
\end{table}}

At first, we add the IOP or COP to the default PAA head, and observe that both IOP and COP can improve the baseline with considerable margins. For the Eff-B0, A0, R18, R50 backbones, IOP increases $+1,+0.7,+0.9,+1.2 \rm AP$, and COP increases $+1.1,+0.8,+1.1,+1.2 \rm AP$, respectively. COP and IOP perform equally on R50, but COP is slightly better for small backbones Eff-B0($+0.1 \rm AP$), A0 ($+0.1 \rm AP$), and R18 ($+0.2 \rm AP$). 

Secondly, we try dropping the IoU prediction during inference and use only IOP or COP, and observe that the results remain almost unchanged (\nth{4} and \nth{5} rows).

However, when we train the models without the IoU branch, the performance is dropped more severely for ResNet50 backbone (\nth{6} and \nth{7} rows). This proves that IoU is still helpful as deep supervised signal for the regression branch in training, but can be safely omitted during infer.  

\section{Prediction Error Analysis}
Beyond evaluating the $\rm mAP$ metric, we use the TIDE \cite{tide-eccv2020} toolbox to analyze the prediction errors of the three models, PAA, CoLAD, and CoLAD-COP, with the same backbone ResNet50. 

\begin{figure}[!t]
	\centering
	\subfloat[\centering PAA]{{\includegraphics[width=0.33\linewidth]{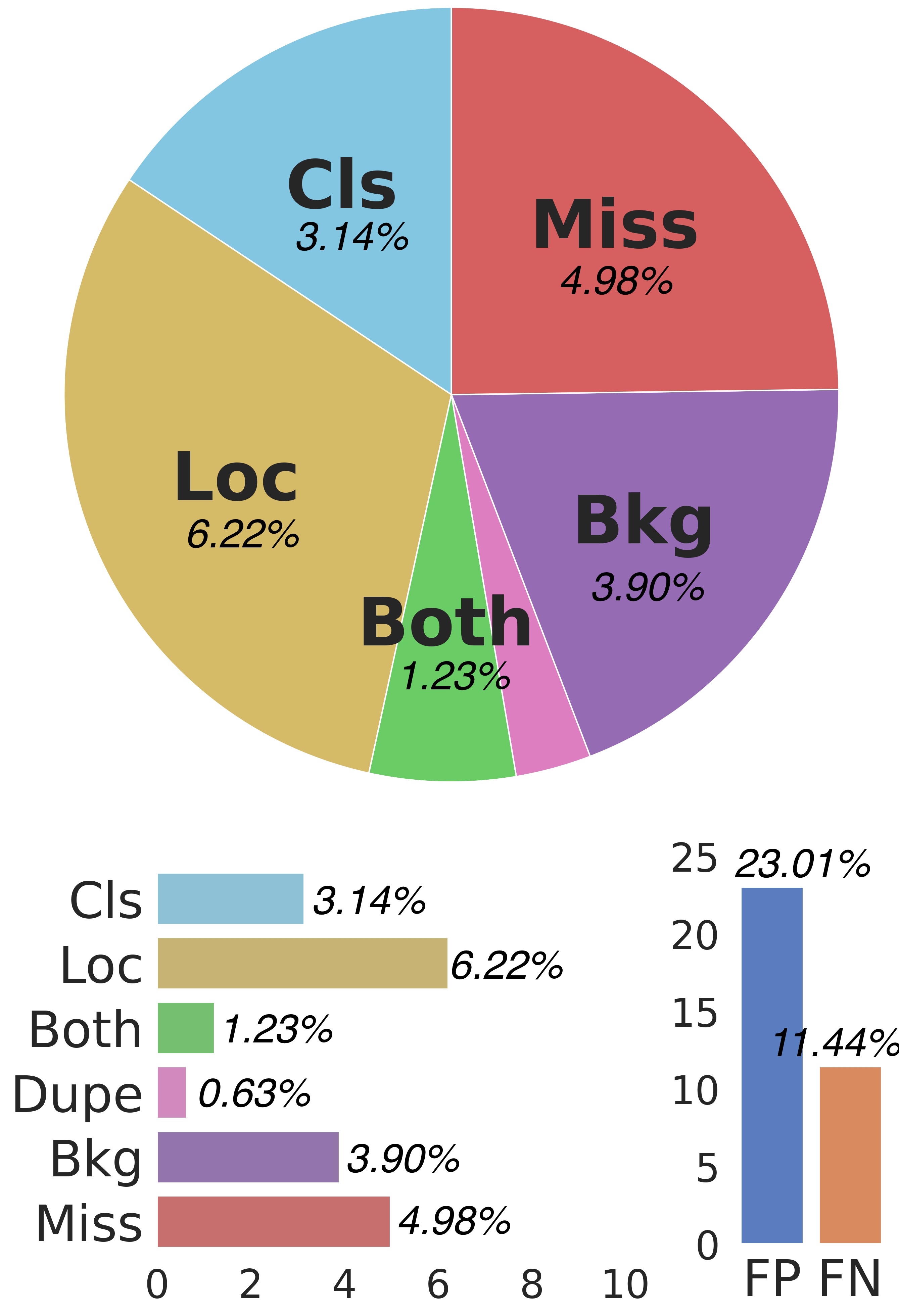} }}
	\subfloat[\centering CoLAD]{{\includegraphics[width=0.33\linewidth]{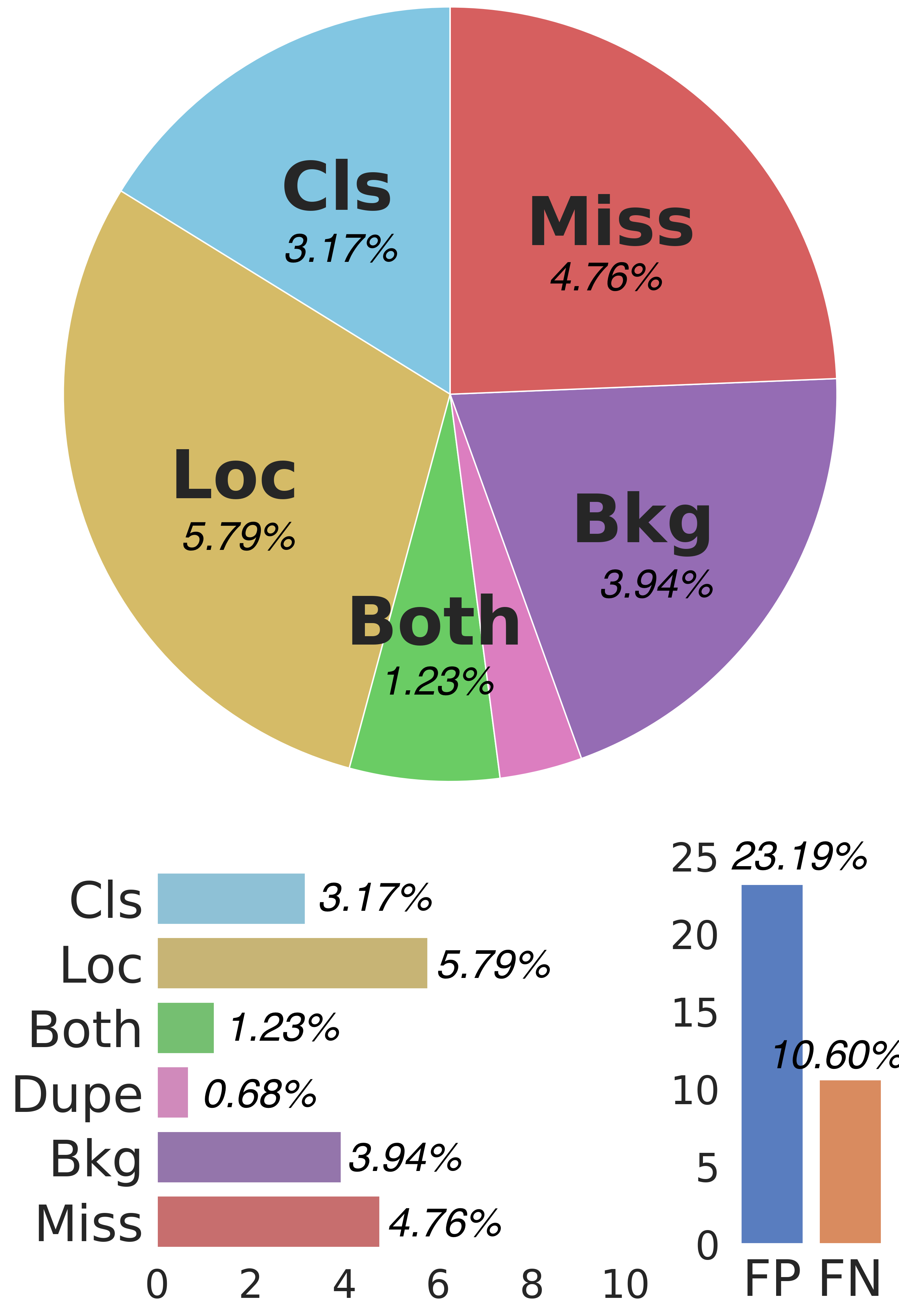} }}
	\subfloat[\centering CoLAD-COP]{{\includegraphics[width=0.33\linewidth]{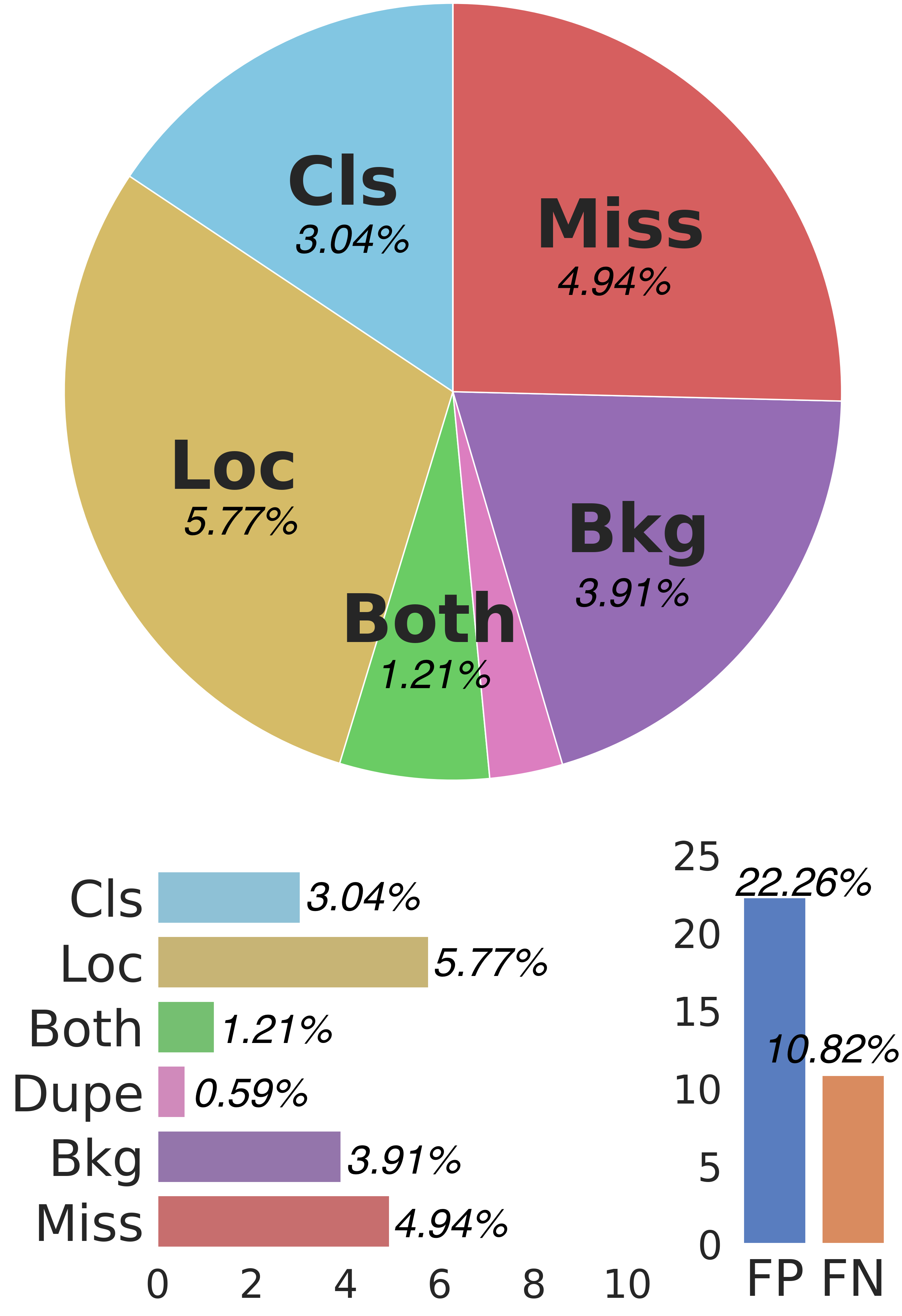} }}
	\caption{\small Error analysis using TIDE \cite{tide-eccv2020} toolbox of the models PAA, CoLAD, and CoLAD-COP with the same backbone ResNet50 on the MS COCO \textit{minval} set.}
	\label{fig:tide}
\end{figure}

As shown in \fig{tide}, CoLAD and CoLAD-COP help reduce the localization error of the baseline PAA from $6.22\%$ to $5.79\%$ and $5.77\%$, respectively. CoLAD also reduces the classification error by $3.04\%$. These indicate that the dynamic mechanism in CoLAD is effective to guide the network to learn a good label assignment, which results in low classification and localization error. In addition, CoLAD can recall more objects, since the false negative percentage is reduced from $11.5\%$ for PAA to $10.85\%$ for both CoLAD and CoLAD-COP. Finally, the introduction of COP can better suppress noisy prediction, as the false positive ratio is reduced from $23.01\%$ to $22.26\%$.

\section{Compare with other distillation methods}
Head-to-head comparison of distillation methods for object detection is not easy, since each method is typically developed for a particular detector. Therefore, for reference purpose only, we select LD \cite{zheng2021localization} to compare, since it is based on the SOTA single-stage detector Generalized Focal (GF) \cite{li2020generalized,li2021generalized}, which is inline with us but has higher performance than our PAA baseline. However, we emphasize that the two methods address different problems. LD\cite{zheng2021localization} focuses on localization distillation and is applied particularly for GFL detector, while we address the label assignment. Therefore, the two methods can be combined. 
 
\begin{table}[!h]
	\caption{Compare our LAD techniques to Localization Distillation (LD) \cite{zheng2021localization} for different ResNet backbones. T and S denote teacher and student networks. LAD is based on PAA\cite{kim2020probabilistic} and LD is based on GF\cite{li2020generalized}. The results are compared for student networks \wrt its baseline on COCO \textit{test} set.}
	\label{tab:resnet_backbone_compare}
	\centering
	\begin{adjustbox}{max width=\columnwidth}
		\small{
			\begin{tabular}{|c|c||c|c|c|c||c|c|}
			\hline
			\textbf{T}    & \textbf{S}   & \textbf{PAA}  & \textbf{LAD}  & \textbf{CoLAD} & \textbf{SoLAD} & \textbf{GF}  & \textbf{LD}   \\ \hline
			R50  & R18 & 35.8 & 36.9 & 36.5  & 38    & 36.0 & 36.1 \\ \hline
			R101 & R18 & 35.8 & 36.8 & 36.6  & 38.4  & 35.8 & 36.5 \\ \hline
			R101 & R50 & 40.4 & 41.6 & 41.3  & 42.4  & 40.1 & 41.1 \\ \hline
		\end{tabular}}
	\end{adjustbox}
\end{table}

\Tab{resnet_backbone_compare} compare the two methods using the same teacher and student's backbones. It is obvious that LAD and CoLAD are superior to LD in all cases. Moreover, our LAD/CoLAD is very simple and can be adapted quickly to any single-stage detectors without architecture modification, and not restricted to Generalized Focal detector \cite{li2020generalized,li2021generalized}. This shows how flexible and effective our method is compared to other distillation methods, such as feature mimicking. 

\end{appendices}

\end{document}